\documentclass{article}

\newcommand{\baselineone}{\textsc{Vlp}}
\newcommand{\baselinetwo}{\textsc{Asrr}}
\newcommand{\ours}{\textsc{Coda}}

 \usepackage[preprint]{neurips_2026}

\usepackage{graphicx}   
\usepackage[table]{xcolor} 
\usepackage{colortbl}   
\usepackage{array}      
\usepackage{booktabs}   
\usepackage{multirow}   
\usepackage{multicol} 
\usepackage{microtype}
\usepackage{caption}
\usepackage{makecell}
\usepackage{siunitx}
\usepackage{paralist}
\setcitestyle{numbers,square} 
\usepackage{hyperref}
\hypersetup{
    colorlinks,
    linkcolor={blue!80!black},
    citecolor={blue!80!black},
}
\usepackage{tcolorbox}
\tcbuselibrary{skins,breakable}
\usepackage{fontawesome}

\newcounter{paranum}
\renewcommand{\theparanum}{\arabic{paranum})}
\newcommand{\numparagraph}[1]{%
  \refstepcounter{paranum}%
  \paragraph{\theparanum\enspace #1}%
}

\definecolor{questioncolor}{RGB}{52, 73, 94}
\definecolor{responseonecolor}{HTML}{c0c0c0}
\definecolor{responsetwocolor}{HTML}{5d80fa}
\definecolor{answercolor}{RGB}{44, 62, 80}

\usepackage{xparse}

\NewDocumentCommand{\modelquestion}{O{Question} m}{%
  \begin{tcolorbox}[
    enhanced,
    breakable,
    colback=white!95!questioncolor,
    colframe=questioncolor,
    title=#1,
    fonttitle=\bfseries,
    arc=1mm,
    boxrule=0.5pt,
    width=\linewidth
  ]
    #2
  \end{tcolorbox}%
}

\newcommand{\modelgt}[1]{%
    \begin{tcolorbox}[
    colback=white!95!answercolor,
    colframe=answercolor,
    title=Ground Truth,
    fonttitle=\bfseries
    ]
      #1
    \end{tcolorbox}
}
\newcommand{\modelresponseone}[1]{%
  \begin{tcolorbox}[
    enhanced,
    breakable,
    colback=white!98!responseonecolor,
    colframe=responseonecolor,
    title=Model Response (GRPO),
    fonttitle=\bfseries,
    arc=1mm,
    boxrule=0.5pt,
    width=\linewidth
  ]
    #1
  \end{tcolorbox}%
}

\newcommand{\modelresponsetwo}[1]{%
  \begin{tcolorbox}[
    enhanced,
    breakable,
    colback=white!98!responsetwocolor,
    colframe=responsetwocolor,
    title=Model Response (\ours{}),
    fonttitle=\bfseries,
    arc=1mm,
    boxrule=0.5pt,
    width=\linewidth
  ]
    #1
  \end{tcolorbox}%
}

\usepackage[utf8]{inputenc} 
\usepackage[T1]{fontenc}    
\usepackage{hyperref}       
\usepackage{url}            
\usepackage{booktabs}       
\usepackage{amsfonts}       
\usepackage{nicefrac}       
\usepackage{microtype}      
\usepackage{xcolor}         

\usepackage{bbm}
\usepackage{amssymb}
\usepackage{amsmath}
\usepackage{subcaption}
\usepackage{amsmath, amssymb}
\usepackage{amsthm}

\theoremstyle{definition}

\title{\ours{}: Difficulty-Aware Compute Allocation\\ for Adaptive Reasoning}

\author{
    Siye Wu\textsuperscript{\rm $\spadesuit$} \quad
    Jian Xie\textsuperscript{\rm $\spadesuit$} \quad 
    Yikai Zhang\textsuperscript{\rm $\spadesuit$} \quad
    Yanghua Xiao\textsuperscript{\rm $\spadesuit$}\\
\textsuperscript{\rm $\spadesuit$}Fudan University\\
{\small \texttt{siyewu24@m.fudan.edu.cn,  shawyh@fudan.edu.cn}}
\\
\\
Project Page: \url{https://coda-site.github.io/} 
}

\begin{document}

\maketitle

\begin{abstract}
The emergence of large reasoning models demonstrates that scaling inference-time compute significantly enhances performance on complex tasks.
However, it often falls into another trap: overthinking simple problems, where repetitive rationales yield minimal accuracy gains at a disproportionately high cost.
This motivates \textit{adaptive reasoning}: dynamically aligning reasoning depth with instance difficulty.
In this paper, we study adaptive reasoning from an optimality perspective, formalizing it as a utility maximization problem where tokens are allocated until the marginal accuracy gain falls below the incremental cost.
Based on this, we propose \ours{} (\textbf{Co}mpute Allocation by \textbf{D}ifficulty \textbf{A}wareness), a method that operationalizes this principle by allocating tokens via a policy-internal difficulty signal.
Specifically, \ours{} estimates difficulty via group-based rollouts and maps it to two non-negative gates that modulate a length-dependent shaping term on top of the binary base reward.
The easy-side gate penalizes verbosity on simple instances, whereas the hard-side gate encourages more deliberative rollouts on challenging ones.
Across model scales and benchmarks, \ours{} achieves adaptive reasoning without external annotations or user-provided budgets:
on easy tasks, \ours{} reduces token costs by over \(60\%\) while maintaining strong accuracy, whereas on hard tasks it incentivizes more deliberative rollouts to maximize performance.

\end{abstract}

\section{Introduction}
\label{sec: introduction}
Recent progress in large language models (LLMs) highlights the value of test-time scaling for reasoning~\citep{liu2025deepseek, openai2025gpt5p2, huang2026step, google2026gemini31pro}.
Specifically, Reinforcement Learning with Verifiable Rewards (RLVR) has facilitated the emergence of large reasoning models (LRMs) by leveraging simple rule-based rewards through algorithms like Group Relative Policy Optimization (GRPO)~\citep{shao2024deepseekmath}, effectively eliciting Long Chain-of-Thought (Long CoT) reasoning~\citep{zeng2025simplerlzoo, hu2025openreasonerzero}.
However, longer reasoning also increases inference cost and can induce overthinking.
On many easy instances, models tend to produce verbose narratives that include redundant steps with little or no gain in correctness~\citep{wu2024how, sui2025stop, chen2025do}.
This mismatch between compute and utility creates a practical bottleneck for deploying reasoning models at scale~\citep{li2025system}.

\begin{figure}[t]
  \centering
  \begin{subfigure}[t]{0.7\linewidth}
    \centering
    \includegraphics[width=\linewidth]{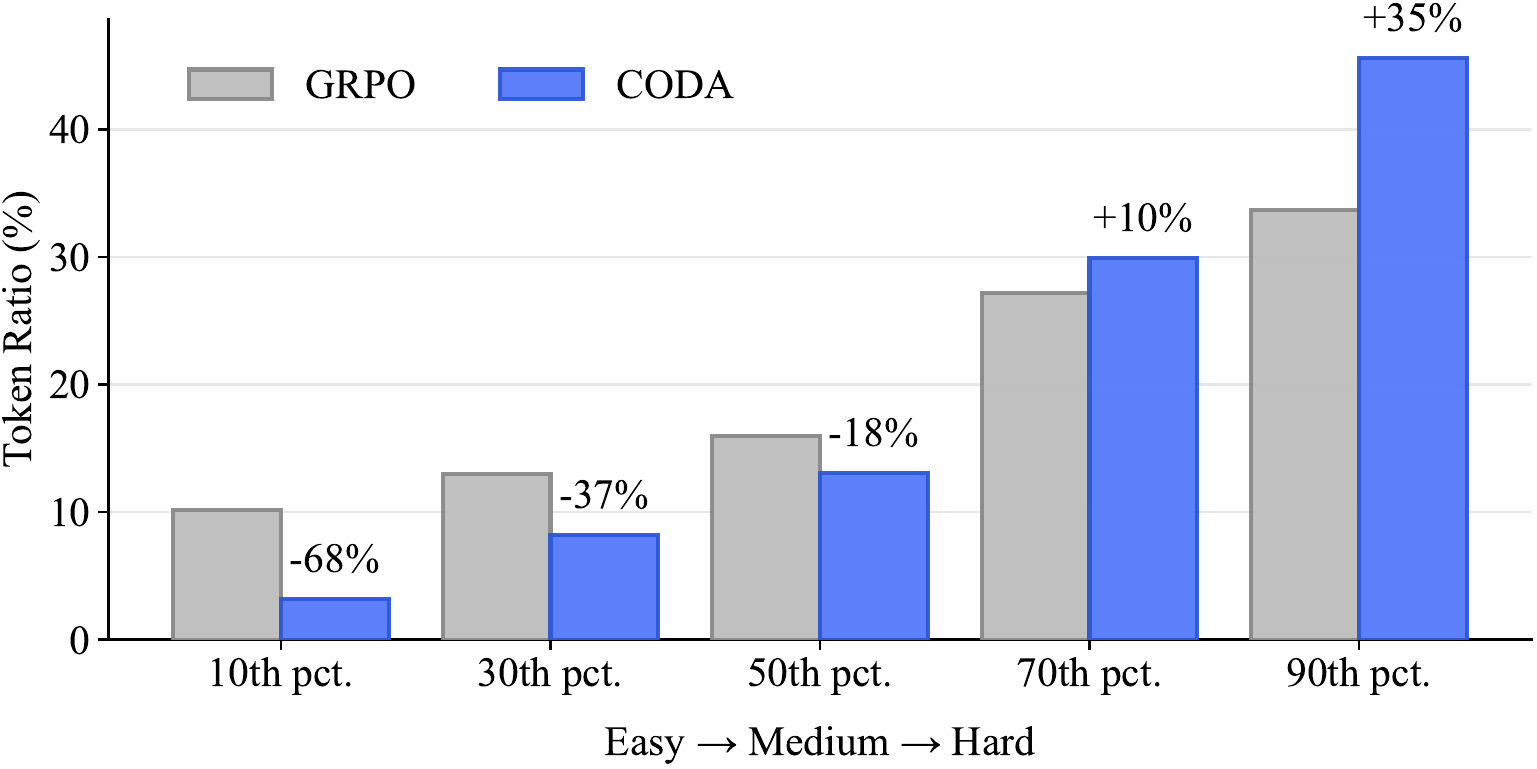}
    \caption{\ours{} allocates fewer tokens to easy tasks and more to hard tasks.}
    \label{fig: main1}
  \end{subfigure}
  \hfill
  \begin{subfigure}[t]{0.28\linewidth}
    \centering
    \includegraphics[width=\linewidth]{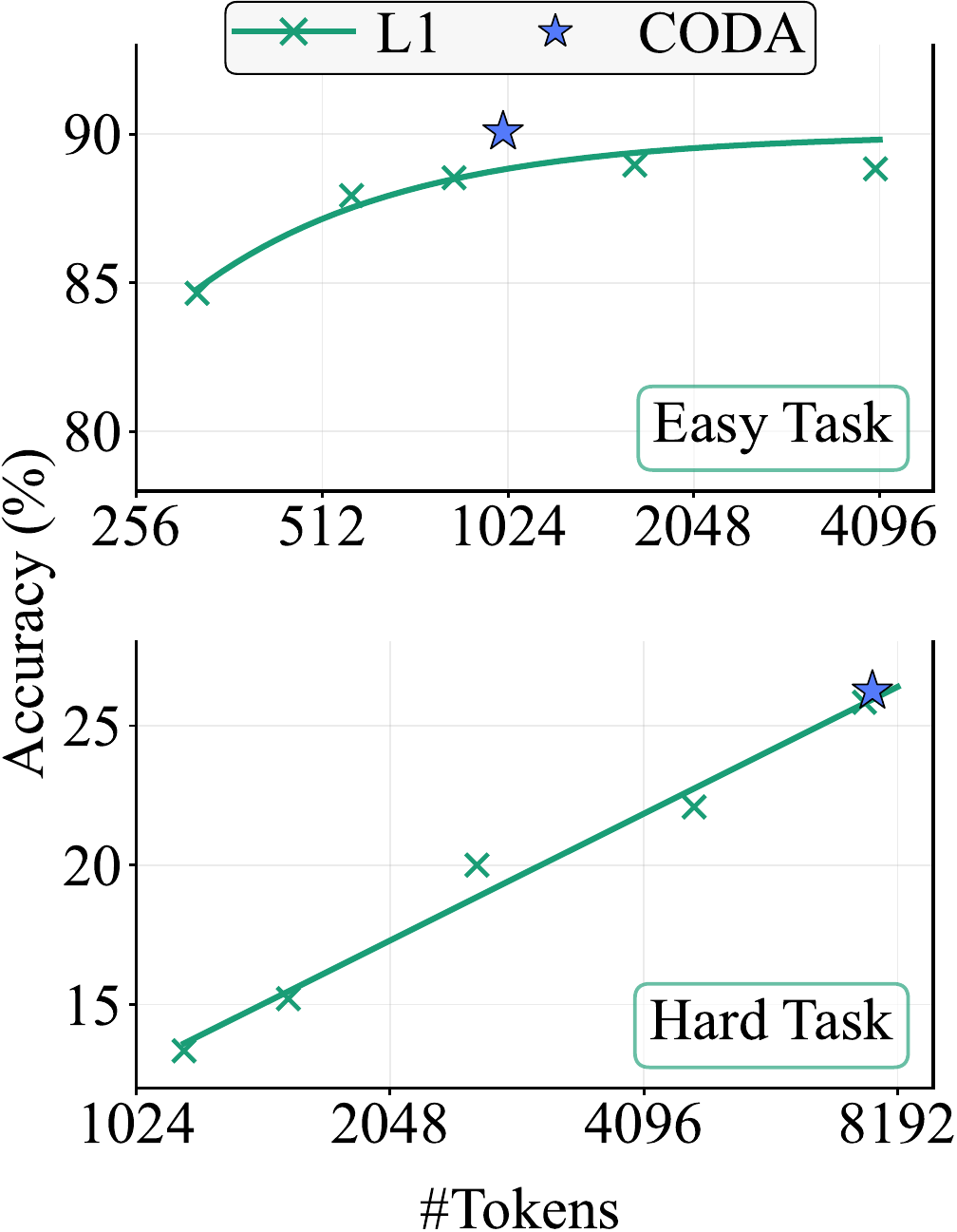}
    \caption{Accuracy vs. \#Tokens.}
    \label{fig: main2}
  \end{subfigure}
  \caption{Adaptive compute allocation across difficulty levels on Qwen3-8B-Base. \textbf{(a)} Compared to GRPO, \ours{} dynamically allocates reasoning tokens by question difficulty, consuming substantially fewer tokens on easier problems while increasing compute for harder ones. \textbf{(b)} On easy tasks (GSM8K, MATH), extra tokens yield marginal gains, and \ours{} achieves optimal accuracy at minimal cost by avoiding unnecessary reasoning. On hard tasks (AIME24\&25), additional tokens substantially improve performance, and \ours{} encourages deeper reasoning to maximize accuracy. 
  }   
  \vspace{-1em}
  \label{fig: main}
\end{figure}

A natural idea is to penalize length during training, which can reduce token usage but often hurts accuracy on hard instances requiring deep reasoning~\citep{xia2025tokenskip, xiang2025just, arora2025training}.
Alternatively, methods like L1~\citep{aggarwal2025l} allow users to specify a target budget during inference.
As Figure~\ref{fig: main2} shows, this provides direct control but is often sensitive to difficulty: underestimating the budget can hurt performance, while overestimating it leads to wasted compute~\citep{hou2025thinkprune, wu2025arm}.
Broadly, many approaches described as adaptive reasoning trade compute for accuracy by deciding when to think more, yet their efficiency gains can still come from accepting modest performance drops for cost savings over typical inputs~\citep{luo2025adar, tu2025learning, zhang2025adaptthink}.
Together, these limitations motivate correctness-aligned objectives that allocate compute by difficulty.

In this work, we study adaptive compute allocation from an optimality perspective.
Viewing generation length as a controllable budget, the optimal policy should spend tokens until the marginal accuracy gain falls below its cost.
Crucially, this balance point depends on difficulty: as shown in Figure~\ref{fig: main2}, easy tasks saturate quickly so extra tokens yield marginal gains, whereas hard tasks continue to improve with more compute.
This suggests that adaptive reasoning should allocate compute by instance difficulty, rather than uniformly shortening or lengthening generations.

Motivated by this principle, we propose \ours{} (\textbf{Co}mpute Allocation by \textbf{D}ifficulty \textbf{A}wareness), which operationalizes difficulty-aware token pricing using a policy-dependent signal.
Specifically, \ours{} estimates difficulty via the group success rate \(s_q\): higher \(s_q\) indicates easier instances for the current policy, and lower \(s_q\) indicates harder ones.
This signal is mapped to two non-negative gates, \(w_q^{\text{easy}}\) and \(w_q^{\text{hard}}\), which scale a length-dependent shaping term on the binary base reward. 
Specifically, \(w_q^{\text{easy}}\) increases the penalty for lengthy outputs when an instance appears easy, discouraging unnecessary verbosity, whereas \(w_q^{\text{hard}}\) adds a bonus for more deliberative rollouts when an instance appears hard, encouraging deeper reasoning when it is likely to help.
As shown in Figure~\ref{fig: main}, \ours{} learns an adaptive allocation rule, consuming substantially fewer tokens on easier problems while reserving more compute for harder ones, without requiring external annotations or user-specified budgets.

Our results and analysis further reveal that:
\begin{inparaenum}[\it 1)]
\item
\ours{} is robust to difficulty shifts.
Under extreme easy- or hard-skewed training, the gating dynamics adjust accordingly while preserving the same qualitative behavior: avoiding unhelpful reasoning and allowing deeper reasoning when needed.
\item 
The easy-side penalty acts as a selective regularizer rather than a blunt truncation.
A moderate penalty trims redundancy without sacrificing base reward, whereas an excessive penalty suppresses exploration.
The hard-side bonus must be correctness-aligned, and rewarding length on incorrect outputs induces degenerate length-seeking without improving accuracy.
\item
\ours{} changes reasoning behavior beyond output length control.
On easy questions, it mitigates overthinking by reducing redundant restatement.
On hard questions, it preserves reflective Long CoT, matching GRPO's reflection patterns while avoiding the indiscriminate truncation of length-penalty baselines.
\end{inparaenum}

In summary, we highlight our contributions:
\begin{inparaenum}[\it 1)]
\item \textbf{An optimality formulation for difficulty-aware compute allocation.}
We formulate compute allocation as utility maximization under token costs, showing that difficulty induces different effective token prices and motivating difficulty-conditioned cost weighting of tokens.
\item \textbf{Dual-gated length shaping from rollout-based difficulty.}
We introduce \ours{}, which estimates difficulty via group success rate \(s_q\) and maps it to \(w_q^{\text{easy}}\) and \(w_q^{\text{hard}}\) to modulate a length-dependent shaping term, penalizing verbosity on easy instances and providing a bonus for deliberative, correctness-aligned rollouts on hard ones.
\item \textbf{Evidence for genuine, robust adaptiveness.}
\ours{} exhibits adaptive reasoning: it allocates tokens by difficulty, remains stable under extreme skews, and is supported by ablations and reasoning-pattern analyses.
\end{inparaenum}

\section{Optimality View of Adaptive Compute}
\label{sec: preliminary}
\paragraph{Compute as a controllable resource.}
For autoregressive language models, generation length provides a practical proxy for measuring inference compute.
Let \(n(o)\) denote the number of generated tokens in an output \(o\).
For a question \(q\), consider a \textit{budgeted decoding} scheme that enforces an explicit effective token budget \(n\), e.g., truncating at \(n\) tokens, or conditioning decoding on a target budget~\citep{hou2025thinkprune, aggarwal2025l},
and let \(\Pr(\text{correct}\mid q,n)\) denote the resulting success probability.
We measure compute by an increasing cost function \(C(n)\), which is typically proportional to token count, and define the utility of allocating \(n\) tokens to \(q\):
{\footnotesize
\begin{equation}
U_q(n)\ \triangleq\ \Pr(\text{correct}\mid q,n)\ -\ \lambda\, C(n),
\label{eq:prelim_utility}
\end{equation}
}
where \(\lambda \ge 0\) controls the cost--accuracy trade-off.
Empirically, \(\Pr(\text{correct}\mid q,n)\) often shows fast saturation on easier questions, whereas additional tokens can still yield substantial gains on harder ones, as shown in Figure~\ref{fig: main2}.

\paragraph{Difficulty and practical proxies.}
Let \(d_q\) denote the difficulty of question \(q\) under the current policy, where a larger \(d_q\) indicates a harder instance.
Since \(d_q\) is not directly observable, previous work often relies on human priors or dataset-specific heuristics~\citep{shi2025efficient, lou2025adacot}.
Importantly, \(d_q\) is policy-dependent: as the policy improves, the same instance may become ``easier''.
In the following analysis, we treat \(d_q\) as an abstract variable and only assume access to a difficulty score that ranks questions consistently with \(d_q\), while we defer the concrete construction of such a proxy to Section~\ref{sec: method}.

\paragraph{Marginal optimality and effective token prices.}
To characterize when extra compute is worthwhile, we study how utility changes with the budget.
Although \(n\) is a discrete integer token count, we quantify the benefit of adding a few more tokens by treating \(n\) as a smooth budget variable and using the slope of \(\Pr(\text{correct}\mid q,n)\) with respect to \(n\), which we approximate by a small difference between nearby budgets.
We then define the marginal gain as
\(g_q(n)\triangleq \frac{\partial}{\partial n}\Pr(\text{correct}\mid q,n)\).
Differentiating Eq.~\eqref{eq:prelim_utility} yields the marginal optimality condition:
{\footnotesize
\begin{equation}
g_q(n)\ \approx\ \lambda\, C'(n),
\label{eq:prelim_marginal}
\end{equation}
}
which balances marginal correctness gain against marginal compute cost.
Equivalently, a small budget increase \(\mathrm{d}n\) changes the expected utility by
\((g_q(n)-\lambda C'(n))\,\mathrm{d}n\), so it is beneficial to increase \(n\) while \(g_q(n)>\lambda C'(n)\) and to stop once \(g_q(n)\le \lambda C'(n)\).
In practice, the marginal gain \(g_q(n)\) decays much faster on easy questions, but can remain sizable over a longer range on hard ones.
This view suggests interpreting \(\lambda\) as a dynamic \textit{effective token price}: a larger \(\lambda\) makes tokens more expensive and induces earlier stopping, while a smaller \(\lambda\) encourages spending more compute when additional reasoning is productive.

\paragraph{Difficulty-aware compute allocation.}
We now connect this optimality view to question difficulty.
Assume a mild structural property: for any fixed budget \(n\), harder questions admit no smaller marginal gains from extra compute, i.e., \(g_q(n)\) is non-decreasing in difficulty \(d_q\).\footnote{This assumption is not required by our algorithm;
it serves as an explanatory lens that motivates difficulty-aware cost weighting.}
Under typical costs (\(C'(n)>0\) and non-decreasing) and diminishing returns in \(n\) (so \(g_q(n)\) eventually decreases with \(n\)), the balance point \(g_q(n)=\lambda C'(n)\) occurs at a larger budget for questions with larger \(g_q(\cdot)\).
In other words, harder questions naturally justify spending more tokens before marginal gains fall below marginal cost, whereas easy tasks saturate much earlier.

It is therefore convenient to re-parameterize this per-question optimal budget as a \textit{question-dependent token price}.
Specifically, we replace the global \(\lambda\) with a multiplier \(\lambda_q\) and consider
\(\Pr(\text{correct}\mid q,n)-\lambda_q C(n)\).
A (local) optimum \(n_q^\star\) satisfies the condition \(g_q(n_q^\star)=\lambda_q C'(n_q^\star)\):
increasing \(\lambda_q\) shifts the marginal-cost curve upward and yields a smaller \(n_q^\star\) (earlier stopping), whereas decreasing \(\lambda_q\) yields a larger \(n_q^\star\) (more compute).
Thus, allocating more tokens to harder questions can be equivalently viewed as assigning them a lower effective token price, with smaller \(\lambda_q\).
In contrast, easy questions face a relatively higher price, with larger \(\lambda_q\).

Finally, Section~\ref{sec: method} turns this token-pricing view into a practical training objective by introducing a policy-internal proxy for relative difficulty and using it to modulate the effective compute cost across questions, enabling difficulty-aware compute allocation without external difficulty annotations.

\section{Method}
\label{sec: method}
Following the optimality principles in Section~\ref{sec: preliminary}, we propose \ours{}, which builds on group-based RL and introduces a lightweight difficulty signal from rollouts for difficulty-aware compute allocation.
Concretely, \ours{}
\begin{inparaenum}[\it 1)]
\item 
estimates instance difficulty using the group success rate \(s_q\);
\item 
converts \(s_q\) into an easy-side penalty weight \(w_q^{\text{easy}}\) and a hard-side bonus weight \(w_q^{\text{hard}}\); and
\item
shapes the base reward \(r_i^{\text{base}}\) with a difficulty-gated length term that discourages redundant verbosity on easy instances while encouraging additional productive reasoning on hard ones.
\end{inparaenum}

\paragraph{Preliminaries: group-based RL.}
In group-based RL methods such as GRPO~\citep{shao2024deepseekmath}, the model samples a group of \(G\) outputs \(O = \{o_1, o_2, \dots , o_G\}\) for each question \(q\).
Each output \(o_i\) then receives a binary base reward
\(
r_i^{\text{base}}=\mathbbm{1}\!\left[\hat{y}_i = y_i\right],
\)
where \(\hat{y}_i\) is the final answer extracted from \(o_i\) and \(y_i\) is the ground truth.
The policy is then optimized with a policy gradient objective using token-level advantages computed from the group rewards.
Although this paradigm enhances reasoning with test-time scaling~\citep{zeng2025simplerlzoo, hu2025openreasonerzero}, it often incurs inference overhead due to verbose generations, especially on easy instances~\citep{sui2025stop, chen2025do}.
A naive remedy is to penalize length directly in the reward.
However, coarse length penalties can be brittle and frequently reduce token usage at the expense of accuracy~\citep{arora2025training, aggarwal2025l}.

\paragraph{Group success rate as a difficulty proxy.}
Since instance difficulty is policy-dependent and not directly observed, we use a lightweight proxy derived from group-based rollouts.
Concretely, we compute the group success rate:
{
\footnotesize
\begin{equation}
s_q \triangleq \frac{1}{G}\sum_{j=1}^{G} r_j^{\text{base}} ,
\end{equation}
}
which serves as an internal difficulty signal: larger \(s_q\) indicates that \(q\) is easier for the current policy, while smaller \(s_q\) indicates a harder question.
Importantly, \(s_q\) requires no external difficulty annotations and naturally evolves as the policy improves during training~\citep{shi2025efficient, lou2025adacot}.

\paragraph{Mapping difficulty to gating weights.}
Based on Eq.~\eqref{eq:prelim_marginal}, we implement adaptive token pricing by scaling a length-dependent shaping term with two non-negative gates, \(w_q^{\text{easy}}\) and \(w_q^{\text{hard}}\).
Specifically, \(w_q^{\text{easy}}\) increases the penalty on lengthy outputs when \(s_q\) is high (easy regime), while \(w_q^{\text{hard}}\) encourages deep reasoning when \(s_q\) is low (hard regime).
To instantiate this, we map \(s_q\) to the two gates as:
{
\footnotesize
\begin{equation}
w_q^{\text{easy}}=\Big[\frac{s_q-\tau_{\text{easy}}}{1-\tau_{\text{easy}}}\Big]_+,\qquad
w_q^{\text{hard}}=\Big[\frac{\tau_{\text{hard}}-s_q}{\tau_{\text{hard}}}\Big]_+,
\label{eq:gate_weights}
\end{equation}
}
where \([\cdot]_+=\max(\cdot,0)\).
This construction ensures \(w_q^{\text{easy}}\) increases with \(s_q\) for easy instances, while \(w_q^{\text{hard}}\) increases when \(s_q\) is small for hard instances.
These thresholds satisfy \(\tau_{\text{easy}}>\tau_{\text{hard}}\) and define the easy and hard regimes by determining when each gate becomes active.

\paragraph{Shaping reward for adaptive compute.}
Let \(|o_i|\) denote the number of generated tokens in rollout \(o_i\). To make the length term scale-invariant across prompts and training stages, we use a group-normalized length \(\tilde{|o_i|}\).
In our implementation, we normalize within the group using the group mean and standard deviation with a small \(\epsilon\), and then apply a sigmoid \(\sigma(\cdot)\) to map the length signal into (0,1).
We shape the base reward with a difficulty-gated length factor:
{
\footnotesize
\begin{equation}
r_i
=
r_i^{\text{base}}
\Bigg(
1
+
\underbrace{\big(\beta\, w_q^{\text{hard}}-\alpha\, w_q^{\text{easy}}\big)}_{\text{difficulty gate}}
\cdot
\underbrace{\sigma(\tilde{|o_i|})}_{\text{relative length}}
\Bigg),
\label{eq:difficulty_gate}
\end{equation}
}
where \(\alpha\ge 0\) controls the easy-side length penalty and \(\beta\ge 0\) controls the hard-side length bonus.
This form yields two desired effects:
\begin{inparaenum}[\it 1)]
\item 
\textbf{Trim redundancy on easy instances.} For easy questions with large \(s_q\), \(w_q^{\text{easy}}\) becomes positive and the term \(-\alpha\, w_q^{\text{easy}}\,\sigma(\tilde{|o_i|})\) down-weights unnecessarily lengthy rollouts, encouraging the policy to stop earlier once reasoning becomes unproductive.
\item
\textbf{Encourage deeper reasoning on hard instances without degenerate length seeking.} For hard questions with small \(s_q\), \(w_q^{\text{hard}}\) activates and the term \(+\beta \, w_q^{\text{hard}} \, \sigma(\tilde{|o_i|})\) increases the reward for more deliberative rollouts.
Crucially, because the entire factor multiplies \(r_i^{\text{base}}\), incorrect rollouts (\(r_i^{\text{base}}=0\)) receive zero reward regardless of length; thus extra length is incentivized only when it is aligned with correctness.
\end{inparaenum}

After computing shaped rewards \(\{r_i\}_{i=1}^G\), we form group-relative advantages for policy optimization by subtracting the group mean reward: \(\hat{A}_{i,k}= r_i - \mathrm{mean}(\{r_i\}_{i=1}^G)\), and apply the standard token-level policy gradient updates.
Refer to Appendix~\ref{sec: objective function} for details.

\section{Experiment}
\label{sec: experiment}
\subsection{Experimental Setup}
\label{subsec: experimental setup}

\paragraph{Models and Training Datasets.} 
To assess the effectiveness of \ours{} across different model sizes, we select Qwen3-4B/8B/14B-Base~\citep{yang2025qwen3} as backbone models.
We use \textit{base} backbones, rather than instruction-tuned variants, to isolate the effect of post-training and to more clearly characterize the training dynamics induced by RL from a common pre-training initialization.
We conduct training on DeepScaleR~\citep{deepscaler2025}, a mathematics dataset of 40K problems spanning a wide range of difficulty levels, drawn from AIME (1983--2023), AMC, Omni-Math~\citep{gao2025omnimath}, and STILL~\citep{min2024imitate}.

\paragraph{Baselines.}
We evaluate \ours{} against the following baselines:
\begin{inparaenum}[\it 1)]
\item \textbf{GRPO}~\citep{shao2024deepseekmath}, a widely used group-based RL method for improving reasoning performance, often benefiting from Long CoT and test-time scaling.
\item 
\textbf{Vanilla Length Penalty (\baselineone{})}~\citep{arora2025training}, which introduces length penalties during RL training to avoid overly long generations. 
\item \textbf{Adaptive Self-Recovery Reasoning (\baselinetwo{})}~\citep{zhang2025continue}, an adaptive length-penalty approach for efficient reasoning that modulates the penalty strength based on estimated difficulty, imposing larger penalties on easier questions.
While effective at reducing token usage, these length-penalty baselines may trade accuracy for brevity by discouraging long reasoning even when it is beneficial on hard instances.
Additional details are provided in Appendix~\ref{sec: length penalty baselines}.
\end{inparaenum}

\paragraph{Implementation Details.}
We implement all RL methods using the open-source training framework verl~\cite{sheng2025hybridflow}.
Unless otherwise specified, we train with a batch size of \(128\) and sample a group of \(G=16\) rollouts per prompt, with a maximum generation length of \(16K\) tokens per rollout.
We run training for one epoch, corresponding to \(314\) global update steps, and set the default sampling temperature to \num{1.0}.
All experiments are conducted on \(8\) NVIDIA H800 GPUs.
For fair comparison, we keep the same training and sampling configuration for all baselines.
To trim redundancy on simple tasks while preserving deliberation for complex ones, we set \(\alpha=\beta=0.2\), \(\tau_{\text{easy}}=0.75\), and \(\tau_{\text{hard}}=0.25\) as stable defaults after light validation.

\paragraph{Evaluation.}
To evaluate reasoning performance under both in-domain and out-of-domain distributions, we consider a diverse suite of general and mathematical benchmarks.
For general reasoning, we use CSQA~\citep{talmor2019commonsenseqa} and GPQA~\citep{rein2024gpqa}, which test commonsense and knowledge-intensive question answering, respectively.
For mathematical reasoning, we evaluate on SVAMP~\citep{patel2021nlp}, GSM8K~\citep{cobbe2021training}, MATH~\citep{hendrycks2021measuring}, AMC23~\citep{maa_amc}, AIME24~\citep{AIME24}, and AIME25~\citep{AIME25}, ranging from elementary to competition-level difficulty.
We evaluate model performance using \textit{accuracy} and \textit{response length} to assess both correctness and generation behavior.
To reduce single-sample decoding variance, we report \(\mathrm{mean@8}\) by sampling \num{8} outputs per prompt.\footnote{For the 4B base model, weaker instruction following may inflate token cost through repetitive outputs.}
During inference, we use temperature \(0.6\) and top-p \(0.95\).

\subsection{Main Results}
\label{subsec: main results}

\definecolor{caribbeangreen}{rgb}{0.0, 0.8, 0.6}
\definecolor{reddishcomplement}{rgb}{1.0, 0.2, 0.4}

\newcolumntype{B}{>{\columncolor{cyan!1}}c}
\newcolumntype{G}{>{\columncolor{orange!1}}c}

\begin{table*}[t]
\caption{
Accuracy (\%) and average token cost of different methods across evaluation benchmarks. $^{\dagger}$ denotes in-domain mathematical benchmarks, and $^{\ddagger}$ denotes out-of-domain general benchmarks. ``\#Tokens'' reports the average number of generated tokens. To reduce single-sample variance, we report \(\mathrm{mean@8}\) results by sampling \(8\) outputs per prompt. ``\(\Delta\)’’ compares \ours{} against GRPO, reporting the accuracy change in percentage points and the relative token reduction.
}
\setlength{\tabcolsep}{3pt}
\centering
\resizebox{\linewidth}{!}{
\begin{tabular}{r| *{9}{B}|*{9}{G}}
\toprule
\multicolumn{1}{c}{\multirow{3}{*}{Models}}
& \multicolumn{9}{c}{\textbf{Accuracy (\%) (\(\uparrow\))}}
& \multicolumn{9}{c}{\textbf{\#Tokens (\(\downarrow\))}} \\
\cmidrule(lr){2-10} \cmidrule(lr){11-19}
\multicolumn{1}{l}{} & \multicolumn{6}{c}{Math\(^{\dagger}\)} & \multicolumn{2}{c}{General\(^{\ddagger}\)} & \multicolumn{1}{c}{\multirow{2}{*}{\cellcolor{white}Avg.}}
& \multicolumn{6}{c}{Math\(^{\dagger}\)} & \multicolumn{2}{c}{General\(^{\ddagger}\)} & \multicolumn{1}{c}{\multirow{2}{*}{\cellcolor{white}Avg.}} \\
\cmidrule(lr){2-7} \cmidrule(lr){8-9} \cmidrule(lr){11-16} \cmidrule(lr){17-18}
\multicolumn{1}{l}{}
& \multicolumn{1}{c}{SVAMP}
& \multicolumn{1}{c}{GSM8K}
& \multicolumn{1}{c}{MATH}
& \multicolumn{1}{c}{AMC23}
& \multicolumn{1}{c}{AIME24}
& \multicolumn{1}{c}{AIME25}
& \multicolumn{1}{c}{CSQA}
& \multicolumn{1}{c}{GPQA}
& \multicolumn{1}{c}{\cellcolor{white}{}}
& \multicolumn{1}{c}{SVAMP}
& \multicolumn{1}{c}{GSM8K}
& \multicolumn{1}{c}{MATH}
& \multicolumn{1}{c}{AMC23}
& \multicolumn{1}{c}{AIME24}
& \multicolumn{1}{c}{AIME25}
& \multicolumn{1}{c}{CSQA}
& \multicolumn{1}{c}{GPQA}
& \cellcolor{white}{} \\
\midrule

Qwen3-4B-Base
& 34.8 & 38.3 & 55.2 & 35.9 & 9.2 & 4.2 & 58.4 & 31.2 & 33.4
& 2805 & 1712 & 1280 & 1091 & 2311 & 2008 & 2193 & 1910 & 1913 \\

GRPO
& 92.5 & 90.9 & 85.3 & 67.2 & 25.8 & 19.6 & 77.7 & 42.6 & 62.7
& 803 & 722 & 2774 & 6042 & 9618 & 9391 & 492 & 3182 & 4128 \\

\baselineone{}
& 92.5 & 91.3 & 84.5 & 67.5 & 22.1 & 17.5 & 78.3 & 42.5 & 62.0
& 245 & 405 & 1716 & 4254 & 6715 & 6298 & 118 & 1983 & 2717 \\

\baselinetwo{}
& 93.1 & 92.0 & 84.8 & 66.3 & 22.5 & 17.9 & 78.0 & 39.5 & 61.7
& 350 & 289 & 1464 & 4022 & 6784 & 6315 & 185 & 1984 & 2674 \\

\ours{}
& 92.8 & 90.5 & 85.1 & 69.7 & 25.4 & 20.8 & 78.9 & 42.6 & 63.2
& 104 & 168 & 1914 & 4809 & 8863 & 8650 & 42 & 1656 & 3276 \\[0.8ex]

\(\Delta\)

& {\textcolor{caribbeangreen}{+\num{0.2}}}
& {\textcolor{reddishcomplement}{-\num{0.4}}}
& {\textcolor{reddishcomplement}{-\num{0.2}}}
& {\textcolor{caribbeangreen}{+\num{2.5}}}
& {\textcolor{reddishcomplement}{-\num{0.4}}}
& {\textcolor{caribbeangreen}{+\num{1.3}}}
& {\textcolor{caribbeangreen}{+\num{1.2}}}
& {\textcolor{caribbeangreen}{\num{0}}}
& {\textcolor{caribbeangreen}{+\num{0.5}}}
& {\textcolor{caribbeangreen}{-\num{87}\%}}
& {\textcolor{caribbeangreen}{-\num{77}\%}}
& {\textcolor{caribbeangreen}{-\num{31}\%}}
& {\textcolor{caribbeangreen}{-\num{20}\%}}
& {\textcolor{caribbeangreen}{-\num{8}\%}}
& {\textcolor{caribbeangreen}{-\num{8}\%}}
& {\textcolor{caribbeangreen}{-\num{91}\%}}
& {\textcolor{caribbeangreen}{-\num{48}\%}}
& {\textcolor{caribbeangreen}{-\num{21}\%}} \\
\midrule

Qwen3-8B-Base
& 81.9 & 81.6 & 67.3 & 47.5 & 11.7 & 9.6 & 74.6 & 36.3 & 51.3
& 239 & 304 & 851 & 1292 & 2317 & 1783 & 252 & 1034 & 1009 \\

GRPO
& 92.9 & 92.5 & 87.8 & 72.2 & 29.6 & 22.1 & 79.8 & 47.1 & 65.5
& 812 & 825 & 2270 & 5022 & 8077 & 7389 & 571 & 3399 & 3546 \\

\baselineone{}
& 92.8 & 92.5 & 87.5 & 70.0 & 25.4 & 20.8 & 79.3 & 47.0 & 64.4
& 312 & 391 & 1801 & 3837 & 7327 & 6247 & 267 & 2002 & 2773 \\

\baselinetwo{}
& 92.8 & 92.5 & 86.7 & 69.1 & 26.7 & 18.8 & 79.7 & 46.5 & 64.1
& 272 & 326 & 1598 & 3688 & 6745 & 5950 & 203 & 1814 & 2575 \\

\ours{}
& 92.9 & 92.5 & 87.6 & 72.5 & 29.6 & 22.9 & 79.9 & 47.2 & 65.6
& 203 & 311 & 1700 & 4159 & 8023 & 7275 & 139 & 2047 & 2982 \\[0.8ex]

\(\Delta\)
& {\textcolor{caribbeangreen}{\num{0}}}
& {\textcolor{caribbeangreen}{\num{0}}}
& {\textcolor{reddishcomplement}{-\num{0.1}}}
& {\textcolor{caribbeangreen}{+\num{0.3}}}
& {\textcolor{caribbeangreen}{\num{0}}}
& {\textcolor{caribbeangreen}{+\num{0.8}}}
& {\textcolor{caribbeangreen}{+\num{0.1}}}
& {\textcolor{caribbeangreen}{+\num{0.1}}}
& {\textcolor{caribbeangreen}{+\num{0.1}}} 
& {\textcolor{caribbeangreen}{-\num{75}\%}}
& {\textcolor{caribbeangreen}{-\num{62}\%}}
& {\textcolor{caribbeangreen}{-\num{25}\%}}
& {\textcolor{caribbeangreen}{-\num{17}\%}}
& {\textcolor{caribbeangreen}{-\num{1}\%}}
& {\textcolor{caribbeangreen}{-\num{2}\%}}
& {\textcolor{caribbeangreen}{-\num{76}\%}}
& {\textcolor{caribbeangreen}{-\num{40}\%}}
& {\textcolor{caribbeangreen}{-\num{16}\%}} \\
\midrule

Qwen3-14B-Base
& 87.1 & 85.8 & 72.5 & 55.6 & 13.3 & 10.4 & 78.8 & 43.7 & 55.9
& 361 & 391 & 914 & 1223 & 2760 & 2610 & 214 & 841 & 1164 \\

GRPO
& 94.1 & 94.0 & 90.3 & 81.6 & 40.0 & 30.4 & 80.7 & 50.6 & 70.2
& 699 & 736 & 2229 & 4522 & 8421 & 8279 & 496 & 2959 & 3543 \\

\baselineone{}
& 94.0 & 93.7 & 90.4 & 78.1 & 37.1 & 27.9 & 81.3 & 50.4 & 69.1
& 264 & 327 & 1821 & 4129 & 7780 & 7546 & 220 & 1981 & 3009 \\

\baselinetwo{}
& 94.0 & 93.6 & 89.1 & 77.5 & 35.0 & 25.8 & 81.2 & 48.8 & 68.1
& 176 & 265 & 1324 & 2998 & 7714 & 7659 & 163 & 1202 & 2687 \\

\ours{}
& 94.1 & 94.0 & 91.1 & 82.2 & 40.8 & 30.4 & 81.3 & 50.6 & 70.6
& 187 & 266 & 1605 & 3684 & 8394 & 8050 & 206 & 1756 & 3018 \\[0.8ex]

\(\Delta\)
& {\textcolor{caribbeangreen}{+\num{0.1}}}
& {\textcolor{caribbeangreen}{\num{0}}}
& {\textcolor{caribbeangreen}{+\num{0.7}}}
& {\textcolor{caribbeangreen}{+\num{0.6}}}
& {\textcolor{caribbeangreen}{+\num{0.8}}}
& {\textcolor{caribbeangreen}{\num{0}}}
& {\textcolor{caribbeangreen}{+\num{0.6}}}
& {\textcolor{caribbeangreen}{\num{0}}}
& {\textcolor{caribbeangreen}{+\num{0.4}}}
& {\textcolor{caribbeangreen}{-\num{73}\%}}
& {\textcolor{caribbeangreen}{-\num{64}\%}}
& {\textcolor{caribbeangreen}{-\num{28}\%}}
& {\textcolor{caribbeangreen}{-\num{19}\%}}
& {\textcolor{caribbeangreen}{\num{0}}}
& {\textcolor{caribbeangreen}{-\num{3}\%}}
& {\textcolor{caribbeangreen}{-\num{58}\%}}
& {\textcolor{caribbeangreen}{-\num{41}\%}}
& {\textcolor{caribbeangreen}{-\num{15}\%}} \\

\bottomrule
\end{tabular}
}
\label{tab: main_results}
\end{table*}

We report our results in Table~\ref{tab: main_results}, and we have the following findings:

\paragraph{\ours{} achieves strong accuracy while substantially reducing token cost.}
Relative to the base backbone, all RL-trained variants improve accuracy, confirming the effectiveness of RL fine-tuning on this setup.
Across all model scales, \ours{} consistently matches or outperforms GRPO in average accuracy while substantially reducing token cost.
For example, with the 8B backbone, \ours{} attains the highest overall accuracy (65.6\%), outperforming all baselines.
Meanwhile, it reduces the average token cost by 16\% compared to GRPO (2982 vs.\ 3546 tokens), primarily by cutting redundant tokens on easier tasks while preserving compute on harder ones.
Similar trends hold for the 4B and 14B models, where \ours{} achieves the best average accuracy with fewer tokens than GRPO.
Moreover, these gains are consistent on both in-domain mathematical benchmarks and out-of-domain general reasoning tasks, suggesting robust generalization beyond the training distribution.
Overall, \ours{} improves efficiency without compromising accuracy, yielding lower token costs at comparable or better performance than existing baselines.

\paragraph{\ours{} effectively allocates compute across difficulty levels.}
A closer look at individual benchmarks shows that \ours{} dynamically allocates compute based on task difficulty.
In contrast to GRPO, which relies on Long CoT even for easy questions, \ours{} reduces unnecessary verbosity by producing shorter generations when lengthier reasoning is unlikely to help.
For instance, with the 8B backbone on SVAMP, \ours{} cuts the average token cost from 812 to 203 (a 75\% reduction) while maintaining comparable accuracy, achieving the lowest token cost among the compared methods.
Conversely, on more challenging benchmarks such as AIME24 and AIME25, \ours{} preserves comparable or higher accuracy with a similar token budget.
Notably, \ours{} does not uniformly shorten generations; instead, it preserves, and can even increase the token budget on hard questions when additional reasoning is needed.
Overall, this behavior reflects adaptive reasoning: saving tokens on easy problems while allocating sufficient reasoning depth to difficult ones.
We further analyze this adaptive compute-allocation behavior in Section~\ref{subsec:adaptiveness_from_internal_signals}.

\paragraph{Compared to length-penalty-based methods, \ours{} improves efficiency without compromising accuracy.}
While \baselineone{} and \baselinetwo{} reduce token usage by penalizing longer generations, they primarily focus on compressing output length rather than adaptively reallocating token budget across varying difficulty levels.
Consequently, their efficiency gains can come from discouraging the complex reasoning trajectories necessary to solve challenging questions, thereby leading to an accuracy-efficiency trade-off.
For example, on AIME25 with the 8B backbone, \baselinetwo{} reduces the average token cost from 7389 to 5950, but its accuracy drops from 22.1\% to 18.8\%.
This performance degradation suggests that adaptive reasoning involves more than simply applying length penalties to shorten generations.
Instead, the policy must autonomously decide when additional tokens are warranted and when to save them to optimize utility.
In contrast, \ours{} dynamically conditions length shaping on estimated difficulty signals, yielding selective efficiency gains on easy problems while preserving sufficient compute to maintain strong performance on hard ones.

\section{Analysis}
\label{sec: analysis}
\subsection{Adaptiveness from Internal Signals} 
\label{subsec:adaptiveness_from_internal_signals}

To verify that \ours{} exhibits genuine adaptive reasoning---reallocating compute with question difficulty rather than indiscriminately shortening or lengthening generations---we conduct additional experiments on 8B models.
Specifically, we compare \ours{} with GRPO on AIME24\&25, MATH500, GSM8K, and CSQA.
We estimate instance difficulty by sampling \(8\) generations per prompt and computing the empirical success rate. 
We then sort instances by this rate, group them into percentile buckets (\(10/30/50/70/90\%\)), and report the token ratio within each bucket.
As shown in Figure~\ref{fig: main1}, \ours{} displays a clear \textit{compute reallocation} pattern: it substantially reduces token usage on easier buckets while allocating more tokens to harder ones.
This trend suggests that \ours{} does not merely enforce shorter outputs.
Instead, it selectively saves tokens when additional reasoning is unnecessary, and reserves compute for instances that benefit from deeper reasoning.

To further contextualize this behavior, we categorize the benchmarks as \textit{Easy} tasks (GSM8K and MATH500) and \textit{Hard} tasks (AIME24\&25), and compare \ours{} with L1~\citep{aggarwal2025l}, a baseline that requires users to explicitly specify token budgets in task instructions.
Refer to Appendix~\ref{sec:l1} for details on L1.
Figure~\ref{fig: main2} shows that this reallocation matches task-specific scaling behavior.
On easy tasks, accuracy saturates quickly as the token budget increases, so additional tokens bring only marginal gains.
In this regime, \ours{} reaches optimal performance with relatively fewer tokens by avoiding redundant reasoning,  thereby improving inference efficiency.
On hard tasks, however, accuracy improves substantially with more compute.
Accordingly, \ours{} allocates more tokens to hard tasks, preserving sufficient reasoning depth when it matters without requiring manual token control.

\subsection{Robustness to Training Difficulty Shifts}
\label{subsec:robustness_to_training_difficulty_shifts}

\begin{table}[t]
  \centering
  \begin{minipage}{0.48\linewidth}
    \centering
    \includegraphics[width=\linewidth]{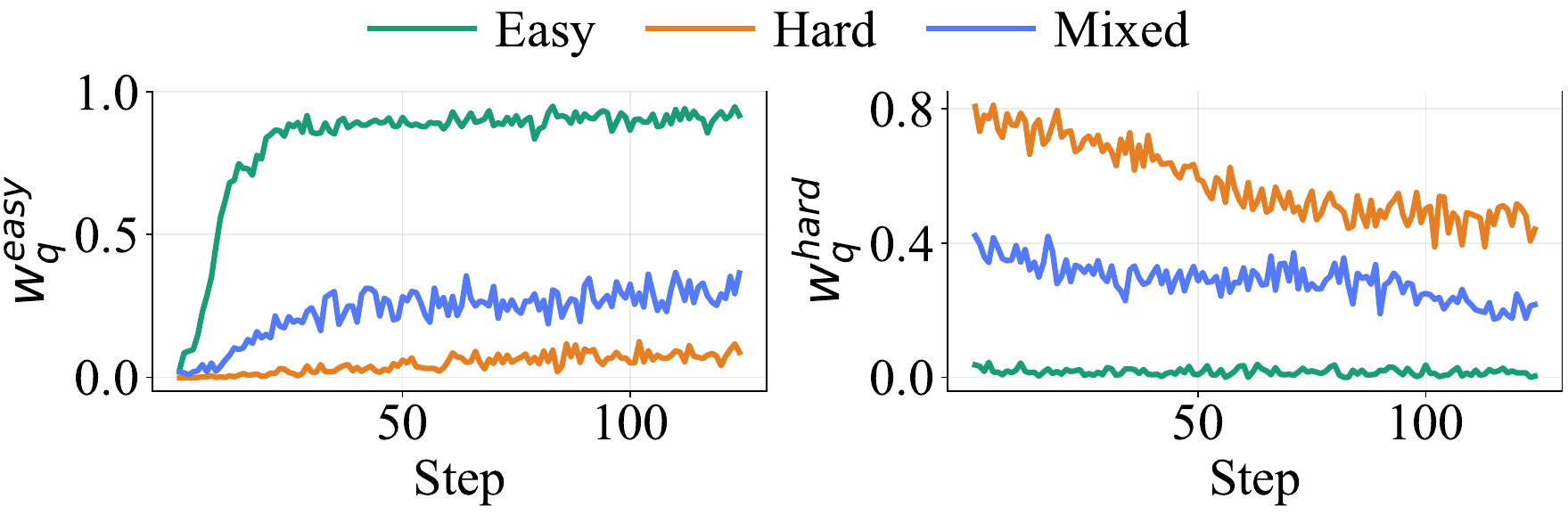}
    \vspace{0pt}
    \captionof{figure}{Dynamics of difficulty-gated weights under different training difficulty distributions, showing that \ours{} automatically adapts its compute allocation to the observed difficulty.
    }
    \label{fig: w dynamics}
  \end{minipage}
  \hfill 
  \begin{minipage}{0.48\linewidth}
    \centering
    \includegraphics[width=\linewidth]{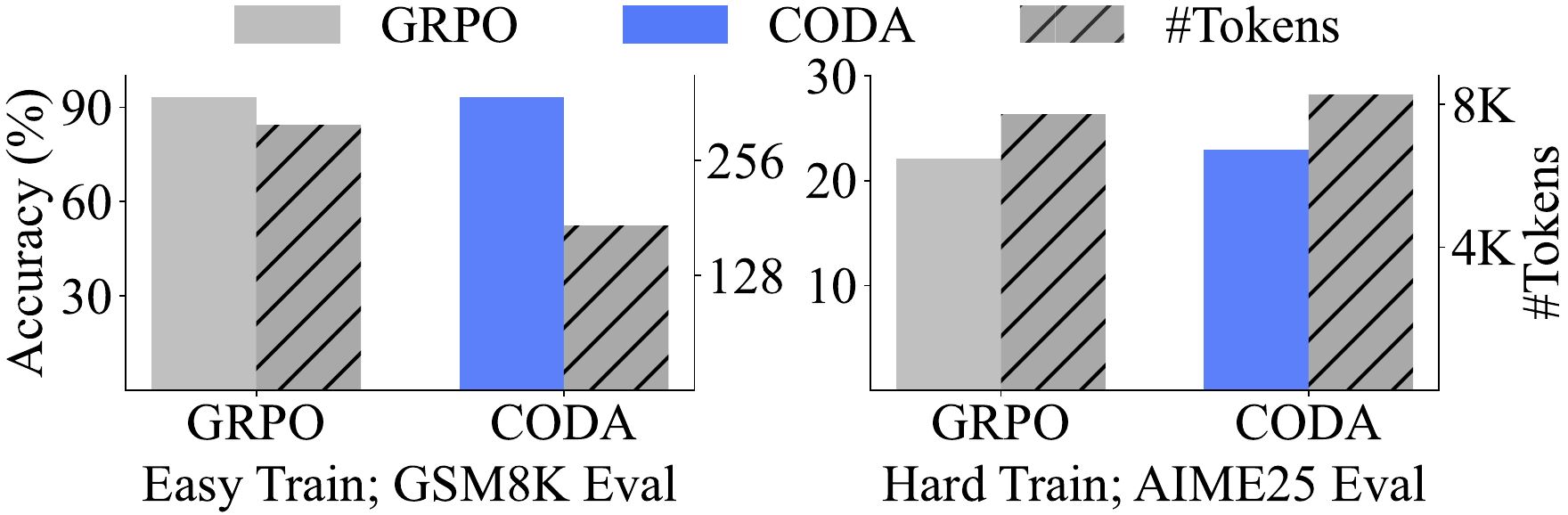}
    \vspace{0pt}
    \captionof{figure}{Robust performance under different training difficulty distributions. \ours{} remains effective across difficulty shifts, maintaining competitive accuracy while adjusting costs.
    }
    \label{fig: easy hard performance}
  \end{minipage}
\vspace{-1em}
\end{table}

To examine whether \ours{}'s inherent adaptiveness remains robust under \textit{extreme} training difficulty skews, we conduct analyses with 8B models under three training settings: \textit{Easy}, \textit{Mixed}, and \textit{Hard}.
We construct separate training datasets for the \textit{Easy} and \textit{Hard} settings as follows.
For \textit{Easy} training, following previous work~\citep{zeng2025simplerlzoo}, we construct an \(\sim\)8K-instance dataset by combining GSM8K with MATH level-1 problems to represent low-complexity reasoning.
For \textit{Hard} training, we adopt a solver-based selection criterion~\citep{shi2025efficient} and sample \(\sim\)13K DeepScaleR problems with \textit{solved percentage} \(< 10\%\).
We provide the complete data processing pipeline in Appendix~\ref{subsec:data_processing}.
We train both the \textit{Easy} and \textit{Hard} settings for 2 epochs.
The \textit{Mixed} setting corresponds to our default setup using the full DeepScaleR training set, as described in Section~\ref{subsec: experimental setup}. 
Our results highlight the following:

\numparagraph{Dynamics of difficulty-gated weights.}
Figure~\ref{fig: w dynamics} shows the training dynamics of the gating signals \(w_q^{\text{easy}}\) and \(w_q^{\text{hard}}\) under different training distributions.
Under \textit{Easy} training, high group success rates \(s_q\) quickly drive \(w_q^{\text{easy}}\) to dominate, while \(w_q^{\text{hard}}\) stays near zero;
conversely, under \textit{Hard} training, persistently low \(s_q\) activates \(w_q^{\text{hard}}\) and keeps \(w_q^{\text{easy}}\) suppressed.
The \textit{Mixed} setting exhibits intermediate behavior, reflecting a balance between the two regimes.
These trends suggest that the gating mechanism is not tied to a particular dataset composition, but instead responds to the difficulty signal perceived by the current policy.
Furthermore, this separation between regimes indicates that the gates track difficulty online and guide policy updates toward an appropriate compute-allocation strategy under the prevailing distribution.
The full training curves are provided in Appendix~\ref{subsec:training_curves}.

\numparagraph{Robust adaptiveness in downstream evaluation.}
To test whether the learned policy transfers beyond the training skew, we further evaluate the resulting models on easy and hard benchmarks (GSM8K, AIME25).
Figure~\ref{fig: easy hard performance} shows that \ours{} remains effective under distribution shifts, despite being trained on heavily skewed data.
After \textit{Easy} training, it achieves competitive GSM8K accuracy at a lower inference cost than GRPO.
In contrast, after \textit{Hard} training, it preserves strong AIME25 performance and correspondingly preserves sufficient depth when additional reasoning is beneficial, rather than collapsing to aggressive truncation.
These results suggest that \ours{} learns a distribution-robust compute allocation rule that reduces unnecessary reasoning when tasks appear easy to the policy while permitting deliberative rollouts when they appear hard.

\setcounter{paranum}{0}

\subsection{Ablations of Difficulty Gating}
\begin{figure*}[t]
\centering
    \includegraphics[width=\linewidth]{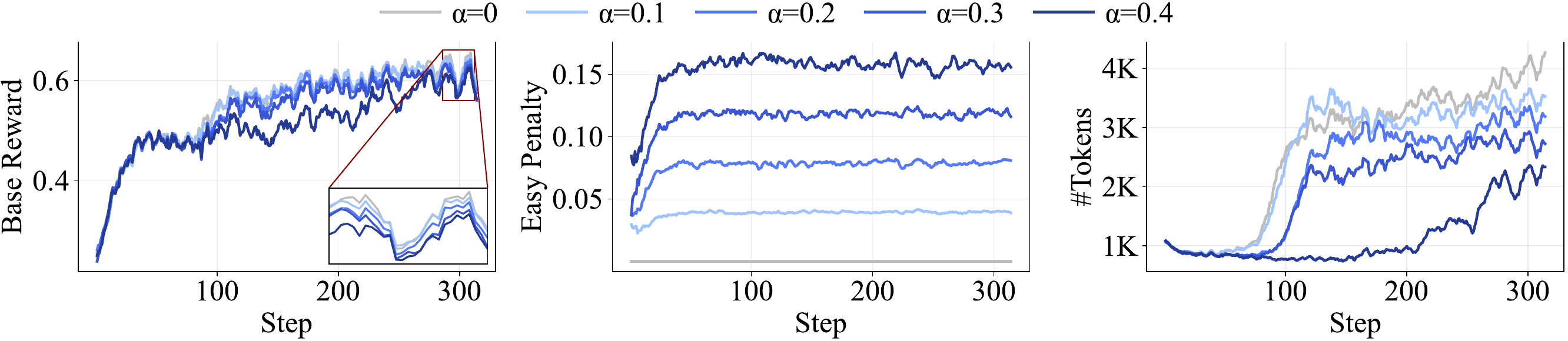}
    \caption{Training dynamics under different easy-penalty strengths \(\alpha\). Moderate \(\alpha\) effectively suppresses unnecessary reasoning while preserving base reward. However, excessively large \(\alpha\) over-penalizes long responses, limiting length scaling and leading to observable gaps in base reward.}
    \label{fig:alpha_dynamics}
    \vspace{-1em}

\end{figure*}

To better understand why \ours{} remains effective, we conduct ablations on 8B models, focusing on two key design choices in our difficulty-gated reward shaping that directly govern the easy–hard compute allocation behavior:
\begin{inparaenum}[\it 1)]
\item the strength of the easy-side length penalty, controlled by \(\alpha\); and
\item whether the hard-side length bonus is aligned with correctness.
\end{inparaenum}

\numparagraph{Effect of easy-penalty strength \(\alpha\).}
We sweep \(\alpha \in [0, 0.4]\) and track the base reward \(r_i^{\text{base}}\), the easy-penalty term \(\alpha\, w_q^{\text{easy}}\, \sigma(\tilde{|o_i|})\), and response length throughout training.
Figure~\ref{fig:alpha_dynamics} shows that a moderate \(\alpha\) substantially reduces token usage while preserving \(r_i^{\text{base}}\), suggesting that the policy learns to stop earlier on easier prompts by trimming redundant steps rather than sacrificing correctness.
In contrast, an overly large \(\alpha\) (e.g., \(\alpha=0.4\)) creates a clear drop in \(r_i^{\text{base}}\) and constrains response length, indicating that an aggressive penalty can begin to hinder exploration and prevent productive long rollouts.
Overall, these results suggest that the easy-side penalty should act as a selective regularizer that trims redundancy, rather than a blunt force that globally truncates generations.
This aligns with the optimality view in Section~\ref{sec: preliminary}: increasing the effective token price on easy instances encourages the policy to stop once additional tokens yield little marginal gain, whereas an overly large \(\alpha\) over-raises the price and starts to suppress beneficial exploration.

\numparagraph{Incorrect-bonus ablation.}
Under our design described in Section~\ref{sec: method}, the length-dependent bonus for hard questions (\(w_q^{\text{hard}} > 0\)) is gated by correctness, so that only correct responses receive additional bonus: \(r_i = 1 + \beta\, w_q^{\text{hard}}\, \sigma(\tilde{|o_i|})\).
To test whether the gains are simply driven by longer outputs (``longer is better''), we also apply the same bonus to incorrect responses.
We then track AIME25 accuracy and response length over training with \(\mathrm{mean@32}\).
Figure~\ref{fig:hard_abla} reveals that rewarding length on incorrect outputs substantially increases response length but does not translate into higher accuracy.
Instead, the policy learns a length-seeking behavior, as length itself becomes a direct optimization target.
This confirms that \ours{} benefits do not come from lengthening per se, but from correctness-aligned incentives that encourage additional reasoning only when it is productive.
In practice, performance is insensitive to \(\beta\) within a reasonable range, while correctness-gating is crucial to prevent degenerate length seeking.
Additional ablations are provided in Appendix~\ref{sec:ablation_studies}.

\subsection{Effects on Reasoning Patterns}
\label{subsec:reasoning_pattern}
To further investigate how \ours{} changes reasoning behavior beyond output length, we analyze its reasoning patterns on 8B models. In particular, we examine two properties implied by our design:
\begin{inparaenum}[\it 1)]
\item on \textit{easy} questions, \ours{} mitigates \textit{overthinking} by avoiding redundant reasoning; and
\item on \textit{hard} questions, \ours{} preserves deep Long CoT reasoning rather than truncating it indiscriminately.
\end{inparaenum}
Additional analysis of \textit{overthinking} for \textit{1)} is provided in Appendix~\ref{sec:overthinking}.

To assess Long CoT reasoning, following previous work~\citep{ma2025rethinking, wu2025arm}, we use \textit{reflective words} as lightweight markers of backtracking and verification.
Concretely, we classify a response as \textit{reflective} if it contains at least one term from a curated list of reflective markers (e.g., \textit{re-evaluate}, \textit{re-think}, and \textit{double-check}); the complete list of \num{17} terms is provided in Appendix~\ref{sec:reflective_words}.
We report \textit{reflection ratio} \(=\mathcal{N}_{\text{ref}}/\mathcal{N}\) and \textit{correct ratio in reflection texts} \(=\mathcal{N}_{\text{ref+}}/\mathcal{N}_{\text{ref}}\), where \(\mathcal{N}\) is the total number of responses, \(\mathcal{N}_{\text{ref}}\) is the number of responses containing at least one reflective marker, and \(\mathcal{N}_{\text{ref+}}\) is the number of correct reflective responses.
Table~\ref{tab:reflective_word} shows the results on AIME24\&25.
Notably, both metrics of \ours{} are comparable to GRPO, suggesting that our easy-side regularization does not suppress reflection on hard problems, but mainly removes redundant reasoning when it is unhelpful.
Moreover, \ours{} consistently outperforms the two length-penalty-based baselines on both metrics, indicating that the preserved reflective Long CoT is not merely longer, but more effective at converting reflection into correctness.
This pattern aligns with our hard-side incentive design, which selectively rewards productive deliberation and thereby maintains the model's capacity for deep, reflective Long CoT reasoning on challenging problems.

\begin{table}[t]
  \centering
  \begin{minipage}{0.64\linewidth}
    \centering
    \includegraphics[width=\linewidth]{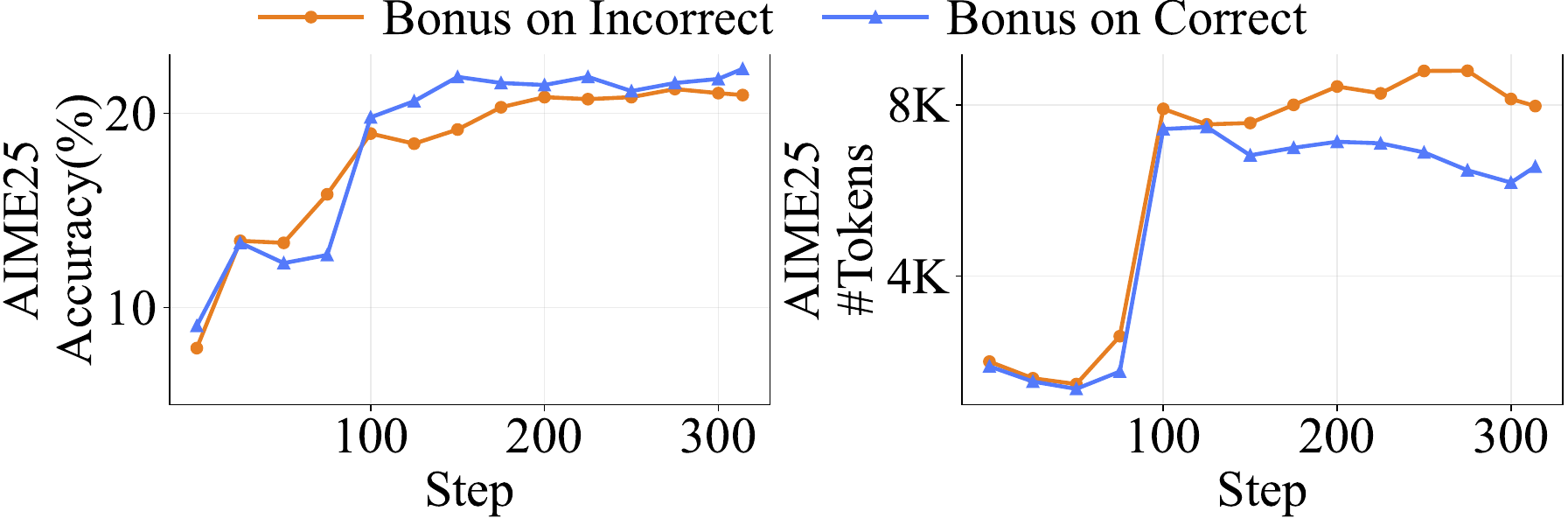}
    \vspace{0pt}
    \captionof{figure}{AIME25 evaluation behavior (\(\mathrm{mean@32}\)) when assigning the length-dependent bonus to correct vs.\ incorrect responses.}
    \label{fig:hard_abla}
  \end{minipage}
  \hfill 
  \begin{minipage}{0.32\linewidth}
    \centering
    \vspace{-20pt}
    \captionof{table}{Results of reflection-related ratios on AIME24\&25.
    }
    \vspace{8pt}
    \resizebox{\linewidth}{!}{
\begin{tabular}{lcc}
\toprule
 & \textbf{\(\frac{\mathcal{N}_{ref}}{\mathcal{N}}\)}\(\uparrow\) & \textbf{\(\frac{\mathcal{N}_{ref+}}{\mathcal{N}_{ref}}\)}\(\uparrow\) \\
\midrule
GRPO   & 94.6\% & 22.7\% \\
\baselineone{}    & 89.8\% & 19.0\% \\
\baselinetwo{}   & 86.9\% & 16.3\% \\
\ours{}   & 93.8\% & 22.9\% \\
\bottomrule
\end{tabular}
}

    \label{tab:reflective_word}
  \end{minipage}
  \vspace{-1em}
\end{table}

\section{Related Work}
\label{sec: related}
\paragraph{Test-time scaling in LLMs.}
Recent advances in LLMs highlight the power of test-time scaling: spending more inference compute, often via longer deliberation traces, substantially improves reasoning performance~\citep{liu2025deepseek, openai2025gpt5p2, huang2026step, google2026gemini31pro}.
In particular, Reinforcement Learning with Verifiable Rewards (RLVR) has attracted growing attention for advancing reasoning capability, contributing to the emergence of large reasoning models (LRMs)~\citep{guo2025deepseek,li2025system}.
With simple rule-based rewards, RLVR algorithms such as Group Relative Policy Optimization (GRPO)~\citep{shao2024deepseekmath} can elicit Long Chain-of-Thought (Long CoT)~\citep{zeng2025simplerlzoo, hu2025openreasonerzero}. This supports deeper reasoning patterns such as searching and self-verification by allocating more inference-time compute~\citep{chen2025towards,zhang2025survey}.
However, this paradigm also incurs substantial overhead from extended outputs, amplifying the ``overthinking'' phenomenon: on easy queries, models often produce redundant restatement and backtracking with little accuracy gain~\citep{wu2024how, sui2025stop, chen2025do}.
This leads to systematic inefficiency and motivates token-efficient compute allocation.

\paragraph{Length control and efficient reasoning.}
A natural approach to efficiency is to control generation length explicitly.
Some methods require or allow users to specify an estimated target budget to guide inference-time compute~\citep{hou2025thinkprune, aggarwal2025l, he2025thinkdial}.
While this offers direct control, it can be sensitive in practice: underestimating the budget can sharply hurt performance, whereas overestimating the difficulty can waste tokens when extra reasoning is unnecessary~\citep{wu2025arm}.
Another line of work reduces token usage by introducing length penalties or constraints during training~\citep{team2025kimi, xia2025tokenskip, xiang2025just, arora2025training}, but such techniques can carry the risk of performance degradation, particularly on challenging instances that genuinely benefit from deep reasoning.
Relatedly, many methods described as \emph{adaptive reasoning} aim to improve the accuracy--efficiency trade-off by learning when to think more or less, yet they frequently achieve efficiency mainly by accepting modest performance drops to unlock large cost savings over the query distribution~\citep{luo2025adar, tu2025learning, zhang2025adaptthink, wu2025arm}.
This may reflect the difficulty of learning a universally optimal trigger, or RL optimization converging to a balance that prioritizes cost savings on typical inputs~\citep{lou2025adacot}.
Overall, these observations motivate difficulty-aware objectives that reallocate compute by trimming redundant verbosity on easy instances while preserving sufficient depth on hard ones, without relying on user-provided budgets or sacrificing correctness as the primary optimization target.

\section{Conclusion}
\label{sec: conclusion}
We study adaptive compute allocation for LLMs through an optimality lens, treating generation length as a controllable inference resource.
This motivates a difficulty-dependent allocation principle: penalizing verbosity on simple instances while encouraging deliberative rollouts on challenging ones.
Building on this insight, we introduce \ours{}, which leverages difficulty signals from group-based rollouts to dynamically modulate reasoning length through a dual-gated reward mechanism.
Across model scales and benchmarks, \ours{} produces genuinely adaptive behavior without external difficulty annotations or user-provided budgets, trimming overthinking on easy inputs---reducing token costs by over \(60\%\)---while preserving effective Long CoT on hard ones.
Further analyses show that \ours{} remains stable under difficulty shifts and that both gating mechanisms are essential for maintaining effectiveness, yielding consistent efficiency gains without sacrificing accuracy on challenging tasks.

\bibliography{src/reference}
\bibliographystyle{plainnat}


\newpage
\appendix

\section*{Appendix}
\label{sec: appendix}
\section{Objective Function}
\label{sec: objective function}

Given a query \( q \) and a set of responses \( O = \{o_1, o_2, \dots, o_{G}\} \) sampled from the old policy \( \pi_{\text{old}} \), we optimize the policy model \( \pi \) using the following objective:
{
\footnotesize
\begin{equation}
\begin{split}
    \mathcal{J}(\theta) = & \mathbb{E}\left[q \sim P(Q), \{o_i\}_{i=1}^{G} \sim \pi_{\theta_{\text{old}}}(O|q)\right] \bigg[ \frac{1}{\sum_{i=1}^{G}|o_{i}|} \sum_{i=1}^{G} \sum_{k=1}^{|o_i|} \Big\{  \min \Big[ \frac{\pi_\theta(o_{i,k} | q, o_{i,<k})}{\pi_{\theta_{\text{old}}}(o_{i,k} | q, o_{i,<k})} \hat{A}_{i,k}, \\
    & \operatorname{clip} \left( \frac{\pi_\theta(o_{i,k} | q, o_{i,<k})}{\pi_{\theta_{\text{old}}}(o_{i,k} | q, o_{i,<k})}, 1 - \epsilon, 1 + \epsilon \right) \hat{A}_{i,k} \Big] 
     - \beta\, \mathrm{KL}\left[\pi_\theta \parallel \pi_{\text{ref}}\right] \Big\} \bigg] ,
\end{split}
\end{equation}
}
where  \( \pi_{\text{ref}} \) denotes the reference model, and the KL divergence term \( \mathrm{KL} \) serves as a constraint to prevent the updated policy from deviating excessively from the reference. 

\section{Details of Length-Penalty Baselines}
\label{sec: length penalty baselines}

\paragraph{\baselineone{}~\citep{arora2025training}.}
\baselineone{} introduces a uniform length penalty to discourage overly long generations.
Specifically, the base reward \(r_i^{\text{base}}\) is multiplicatively modulated by a length-dependent factor:
{
\footnotesize
\begin{equation}
r_i
=
r_i^{\text{base}}
\Bigg(
1
-
\gamma \cdot \sigma(\tilde{|o_i|})
\Bigg),
\end{equation}
}
where \(\sigma(\cdot)\) denotes the sigmoid function and \(\tilde{|o_i|}\) represents the normalized output length.
The coefficient \(\gamma\) controls the overall penalty strength.
Unlike adaptive methods, \baselineone{} applies a uniform penalty regardless of question difficulty.
It does not differentiate between easy and hard problems, and therefore may under-penalize easy samples or over-penalize difficult ones.
Following the original implementation and recommended hyperparameter settings, we set \(\gamma=0.1\).

\paragraph{\baselinetwo{}~\citep{zhang2025continue}.}
\baselinetwo{} introduces an adaptive length penalty.
Rather than applying a uniform regularization, the penalty strength is dynamically modulated by the group-level success rate \(s_q\):
{
\footnotesize
\begin{equation}
r_i
=
r_i^{\text{base}}
-
\zeta
\frac{
\left[ s_q - \tau + \epsilon \right]_+
}{
1 - \tau + \epsilon
}
\cdot
\mathrm{clip}
\left(
\frac{
|o_i| - |o_\text{correct\_shortest}|
}{
L_{\text{window}}
},
\, 0,\, 1
\right).
\end{equation}
}

The penalty is activated only when the estimated difficulty exceeds threshold \(\tau\).
The clipped length term penalizes outputs that exceed the shortest correct solution \(|o_\text{correct\_shortest}|\) within a local window \(L_{\text{window}}\).
The penalty becomes stronger as the group-level success rate increases, thereby discouraging long reasoning on easier questions.
Although \baselinetwo{} adaptively scales the penalty, it focuses on length suppression rather than compute reallocation across difficulty levels, potentially trading accuracy for shorter generations.
Following the original implementation and recommended hyperparameter settings, we set \(\tau=0.75\), \(\zeta=0.5\), \(L_{\text{window}}=2K\), and \(\epsilon=10^{-6}\).

\section{Details of L1}
\label{sec:l1}

L1~\citep{aggarwal2025l} trains a model to follow a user-specified token budget by conditioning the prompt on a target length \(n_{\text{gold}}\).
Concretely, each training prompt is augmented with the instruction ``Think for \(n_{\text{gold}}\) tokens.''
During training, \(n_{\text{gold}}\) is sampled uniformly from \(\mathcal{U}(n_{\text{min}}, n_{\text{max}})\).
As a result, at inference time, L1 requires users to explicitly provide a token budget in the instruction, and the model then attempts to match the target length.
The reward is defined as:
{
\footnotesize
\begin{equation}
r_i = r_i^{\text{base}} - \eta \cdot \bigl|n_{\text{gold}} - |o_i|\bigr|.
\end{equation}
}

Here \(\eta\) controls the trade-offs between correctness and budget adherence. 
Following the original implementation, we set \(\eta = 0.0003\) and use \(n_{\text{min}}=128\) and \(n_{\text{max}}=10K\), respectively.

\section{Details of Training Difficulty Shifts}
\label{sec:data_processing}

\subsection{Data Processing}
\label{subsec:data_processing}

\begin{table*}[t]
\caption{Training data construction for difficulty-shift experiments.}
\label{tab:shift_data}
\centering
    
\begin{tabular}{lcc}
\toprule
\textbf{Dataset} & \textbf{Processing} & \textbf{Size} \\
\midrule
\multicolumn{3}{l}{\textbf{Easy Training}} \\
\midrule
GSM8K~\citep{cobbe2021training} & full & 7.4K \\
MATH~\citep{hendrycks2021measuring} & level-1 & 1.0K \\
\cmidrule(lr){3-3}
& & 8.4K \\
\midrule
\multicolumn{3}{l}{\textbf{Hard Training}} \\
\midrule
DeepScaleR~\citep{deepscaler2025} & \textit{sp} \(< 10\%\) & 13.1K \\
\cmidrule(lr){3-3}
& & 13.1K \\
\bottomrule
\end{tabular}

\end{table*}

To study robustness under \textit{extreme} training difficulty skews, we construct two training distributions for 8B models: \textit{Easy} and \textit{Hard}.
Table~\ref{tab:shift_data} summarizes the resulting datasets.

\paragraph{Easy Training.}
Following prior work~\citep{zeng2025simplerlzoo}, we build an \(\sim\)8K easy-skewed set by combining:
\begin{inparaenum}[\it 1)]
\item the full GSM8K training split (7.4K); and
\item the level-1 subset of MATH (1.0K),
resulting in 8.4K instances in total.
\end{inparaenum}
A sample problem from this split is shown below.

\modelquestion[Question (Easy)]{
Mike decides to buy a new camera.  He wants to buy a top-of-the-line camera but he decides to wait for the new model to come out.  The new model costs 30\% more than the current model.  The old camera cost \$4000.  He gets \$200 off a \$400 lens he bought.  How much did he pay for the camera and lens?
}
\modelgt{
\boxed{5400}
}

\paragraph{Hard Training.}
For the hard-skewed setting, we adopt the solver-based difficulty annotation released by \citet{shi2025efficient}.
Their procedure runs a fixed solver model, Qwen2.5-Math-7B~\citep{yang2024qwen2}, for \(n=128\) independent attempts per problem and records the empirical success rate.
The released field \textit{solved percentage} (\textit{sp}) is defined as \(100 \times (s_i/n)\), where \(s_i\) is the number of successful attempts.
We then select DeepScaleR problems with \textit{sp} \(< 10\%\), yielding 13.1K instances.
A sample problem from this split is shown below.

\modelquestion[Question (Hard)]{
A bee starts flying from point $P_0$. She flies $1$ inch due east to point $P_1$. For $j \ge 1$, once the bee reaches point $P_j$, she turns $30^{\circ}$ counterclockwise and then flies $j+1$ inches straight to point $P_{j+1}$. When the bee reaches $P_{2015},$ how far from $P_0$ is she, in inches?
}
\modelgt{
\boxed{1008 \sqrt{6} + 1008 \sqrt{2}}
}

\subsection{Training Curves}
\label{subsec:training_curves}

\begin{figure*}[t]
\centering
    \includegraphics[width=0.5\linewidth]{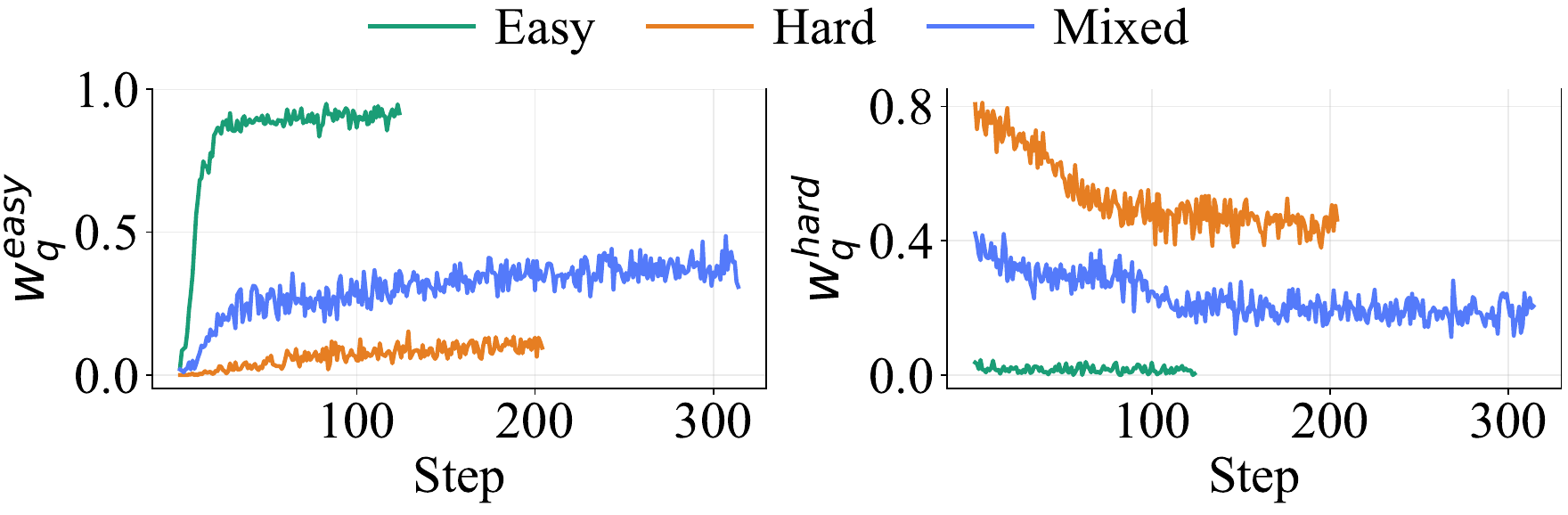}
    \caption{Dynamics of difficulty-gated weights under different training difficulty distributions, showing that \ours{} automatically adapts its compute allocation to the observed difficulty.}
    \label{fig:full_shifts_dynamics}
\end{figure*}

We provide the full training curves for the difficulty-shift experiments in Figure~\ref{fig:full_shifts_dynamics}.

\section{Extended Ablations on Difficulty Gating}
\label{sec:ablation_studies}

\begin{table*}[t]
\caption{
Ablation on easy-side penalty strength \(\alpha\) for \ours{} (8B models).
}
\setlength{\tabcolsep}{3pt}
\centering
\resizebox{\linewidth}{!}{
\begin{tabular}{c| *{9}{B}|*{9}{G}}
\toprule
\multicolumn{1}{c}{\multirow{3}{*}{\(\alpha\)}}
& \multicolumn{9}{c}{\textbf{Accuracy (\%) (\(\uparrow\))}}
& \multicolumn{9}{c}{\textbf{\#Tokens (\(\downarrow\))}} \\
\cmidrule(lr){2-10} \cmidrule(lr){11-19}
\multicolumn{1}{l}{} & \multicolumn{6}{c}{Math\(^{\dagger}\)} & \multicolumn{2}{c}{General\(^{\ddagger}\)} & \multicolumn{1}{c}{\multirow{2}{*}{\cellcolor{white}Avg.}}
& \multicolumn{6}{c}{Math\(^{\dagger}\)} & \multicolumn{2}{c}{General\(^{\ddagger}\)} & \multicolumn{1}{c}{\multirow{2}{*}{\cellcolor{white}Avg.}} \\
\cmidrule(lr){2-7} \cmidrule(lr){8-9} \cmidrule(lr){11-16} \cmidrule(lr){17-18}
\multicolumn{1}{l}{}
& SVAMP & GSM8K & MATH & AMC23 & AIME24 & AIME25 & CSQA & GPQA & \multicolumn{1}{c}{\cellcolor{white}{}}
& SVAMP & GSM8K & MATH & AMC23 & AIME24 & AIME25 & CSQA & GPQA & \\
\midrule

0
& 93.0 & 92.5 & 87.9 & 73.1 & 30.0 & 22.9 & 79.7 & 47.2 & 65.8
& 604 & 655 & 2420 & 5244 & 8635 & 7870 & 412 & 2822 & 3583 \\

0.1
& 93.3 & 92.6 & 87.8 & 72.8 & 29.6 & 22.5 & 79.0 & 46.0 & 65.4
& 558 & 573 & 2003 & 4605 & 8321 & 7328 & 372 & 2832 & 3324 \\

0.2
& 92.9 & 92.5 & 87.6 & 72.5 & 29.6 & 22.9 & 79.9 & 47.2 & 65.6
& 203 & 311 & 1700 & 4159 & 8023 & 7275 & 139 & 2047 & 2982 \\

0.3
& 92.6 & 92.1 & 86.4 & 71.9 & 28.8 & 21.3 & 79.9 & 45.4 & 64.8
& 186 & 231 & 1464 & 3773 & 6939 & 6528 & 86 & 1543 & 2594 \\

0.4
& 91.7 & 90.9 & 84.5 & 67.8 & 27.1 & 19.6 & 78.9 & 44.2 & 63.1
& 191 & 117 & 1081 & 3202 & 6870 & 5872 & 38 & 1236 & 2326 \\

\bottomrule
\end{tabular}
}
\label{tab:alpha_ablation}
\end{table*}
\begin{table*}[t]
\caption{
Ablation on hard-side bonus strength \(\beta\) for \ours{} (8B models).
}
\setlength{\tabcolsep}{3pt}
\centering
\resizebox{\linewidth}{!}{
\begin{tabular}{c| *{9}{B}|*{9}{G}}
\toprule
\multicolumn{1}{c}{\multirow{3}{*}{\(\beta\)}}
& \multicolumn{9}{c}{\textbf{Accuracy (\%) (\(\uparrow\))}}
& \multicolumn{9}{c}{\textbf{\#Tokens (\(\downarrow\))}} \\
\cmidrule(lr){2-10} \cmidrule(lr){11-19}
\multicolumn{1}{l}{} & \multicolumn{6}{c}{Math\(^{\dagger}\)} & \multicolumn{2}{c}{General\(^{\ddagger}\)} & \multicolumn{1}{c}{\multirow{2}{*}{\cellcolor{white}Avg.}}
& \multicolumn{6}{c}{Math\(^{\dagger}\)} & \multicolumn{2}{c}{General\(^{\ddagger}\)} & \multicolumn{1}{c}{\multirow{2}{*}{\cellcolor{white}Avg.}} \\
\cmidrule(lr){2-7} \cmidrule(lr){8-9} \cmidrule(lr){11-16} \cmidrule(lr){17-18}
\multicolumn{1}{l}{}
& SVAMP & GSM8K & MATH & AMC23 & AIME24 & AIME25 & CSQA & GPQA & \multicolumn{1}{c}{\cellcolor{white}{}}
& SVAMP & GSM8K & MATH & AMC23 & AIME24 & AIME25 & CSQA & GPQA & \\
\midrule

0
& 92.7 & 92.5 & 87.3 & 72.2 & 27.5 & 21.3 & 79.7 & 46.1 & 64.9
& 220 & 320 & 1639 & 3522 & 6548 & 5930 & 209 & 1852 & 2530 \\

0.1
& 92.6 & 92.4 & 87.2 & 71.9 & 28.3 & 21.7 & 79.6 & 46.1 & 65.0
& 216 & 313 & 1672 & 3812 & 7581 & 6635 & 198 & 1792 & 2777 \\

0.2
& 92.9 & 92.5 & 87.6 & 72.5 & 29.6 & 22.9 & 79.9 & 47.2 & 65.6
& 203 & 311 & 1700 & 4159 & 8023 & 7275 & 139 & 2047 & 2982 \\

0.3
& 92.7 & 92.4 & 87.7 & 72.5 & 29.2 & 22.5 & 79.5 & 46.8 & 65.4
& 232 & 352 & 1765 & 4157 & 7505 & 6928 & 237 & 1938 & 2889 \\

0.4
& 93.0 & 92.6 & 87.9 & 72.5 & 30.0 & 22.9 & 79.4 & 46.9 & 65.7
& 294 & 351 & 1775 & 4098 & 7929 & 7121 & 219 & 2019 & 2976 \\

\bottomrule
\end{tabular}
}
\label{tab:beta_ablation}
\end{table*}
\begin{table*}[t]
\caption{
Ablation on difficulty-gating thresholds \((\tau_{\text{easy}}, \tau_{\text{hard}})\) for \ours{} (8B models).
}
\setlength{\tabcolsep}{3pt}
\centering
\resizebox{\linewidth}{!}{
\begin{tabular}{c| *{9}{B}|*{9}{G}}
\toprule
\multicolumn{1}{c}{\multirow{3}{*}{\((\tau_{\text{easy}}, \tau_{\text{hard}})\)}}
& \multicolumn{9}{c}{\textbf{Accuracy (\%) (\(\uparrow\))}}
& \multicolumn{9}{c}{\textbf{\#Tokens (\(\downarrow\))}} \\
\cmidrule(lr){2-10} \cmidrule(lr){11-19}
\multicolumn{1}{l}{} & \multicolumn{6}{c}{Math\(^{\dagger}\)} & \multicolumn{2}{c}{General\(^{\ddagger}\)} & \multicolumn{1}{c}{\multirow{2}{*}{\cellcolor{white}Avg.}}
& \multicolumn{6}{c}{Math\(^{\dagger}\)} & \multicolumn{2}{c}{General\(^{\ddagger}\)} & \multicolumn{1}{c}{\multirow{2}{*}{\cellcolor{white}Avg.}} \\
\cmidrule(lr){2-7} \cmidrule(lr){8-9} \cmidrule(lr){11-16} \cmidrule(lr){17-18}
\multicolumn{1}{l}{}
& SVAMP & GSM8K & MATH & AMC23 & AIME24 & AIME25 & CSQA & GPQA & \multicolumn{1}{c}{\cellcolor{white}{}}
& SVAMP & GSM8K & MATH & AMC23 & AIME24 & AIME25 & CSQA & GPQA & \\
\midrule

(0.6, 0.25)
& 92.8 & 92.1 & 87.3 & 72.2 & 29.2 & 22.5 & 79.8 & 46.0 & 65.2
& 183 & 285 & 1621 & 4100 & 7664 & 6956 & 145 & 1797 & 2844 \\

(0.75, 0.25)
& 92.9 & 92.5 & 87.6 & 72.5 & 29.6 & 22.9 & 79.9 & 47.2 & 65.6
& 203 & 311 & 1700 & 4159 & 8023 & 7275 & 139 & 2047 & 2982 \\

(0.9, 0.25)
& 92.8 & 92.6 & 88.1 & 72.5 & 30.0 & 22.9 & 79.5 & 47.0 & 65.7
& 343 & 424 & 1876 & 4372 & 8176 & 7208 & 307 & 2688 & 3174 \\

\midrule

(0.75, 0.1)
& 92.9 & 92.5 & 87.7 & 72.8 & 29.2 & 22.5 & 79.5 & 46.5 & 65.4
& 216 & 312 & 1795 & 4262 & 7787 & 7078 & 248 & 1980 & 2960 \\

(0.75, 0.25)
& 92.9 & 92.5 & 87.6 & 72.5 & 29.6 & 22.9 & 79.9 & 47.2 & 65.6
& 203 & 311 & 1700 & 4159 & 8023 & 7275 & 139 & 2047 & 2982 \\

(0.75, 0.4)
& 93.0 & 92.7 & 87.6 & 72.5 & 30.0 & 22.9 & 79.2 & 46.6 & 65.6
& 261 & 324 & 1892 & 4525 & 8175 & 7856 & 212 & 2021 & 3158 \\

\bottomrule
\end{tabular}
}
\label{tab:gating_ablation}
\end{table*}

Tables~\ref{tab:alpha_ablation},~\ref{tab:beta_ablation}, and~\ref{tab:gating_ablation} provide a more detailed view of the design choices in difficulty-gated reward shaping.
Consistent with our main claims, we observe three key patterns.

\begin{inparaenum}[\it 1)]
\item
Varying the easy-side penalty strength \(\alpha\) reveals a consistent pattern: moderate values substantially reduce token usage while preserving performance, whereas overly large \(\alpha\) over-penalizes long responses and degrades accuracy.
This supports the view that the penalty should act as a selective regularizer rather than a global truncation mechanism.
\item 
Performance is relatively insensitive to the exact value of \(\beta\) within a reasonable range, but becomes unstable when correctness-gating is removed. 
Notably, when \(\beta = 0\), the formulation reduces to a pure length-penalty scheme that trades accuracy for shorter responses, discouraging long outputs without encouraging productive reasoning.
This confirms that the gains of \ours{} arise from correctness-aligned incentives.
\item
The gating thresholds \((\tau_{\text{easy}}, \tau_{\text{hard}})\) show that extreme settings lead to suboptimal compute allocation, indicating that effective adaptiveness requires a balanced separation between easy and hard regimes.

Overall, these results further support that \ours{} achieves adaptive compute allocation by selectively modulating reasoning depth based on difficulty, rather than relying on uniform length control.
\end{inparaenum}

\section{Analysis of Overthinking}
\label{sec:overthinking}

We follow prior studies on the \textit{overthinking} phenomenon~\citep{sui2025stop, chen2025do}:
on relatively simple instances, a model allocates excessive reasoning compute, which often manifests as redundant intermediate steps or repeated verification and backtracking loops, while yielding little or no improvement in correctness.
To further analyze this phenomenon, we compare responses from the GRPO and \ours{} 8B models on representative easy instances from SVAMP.
As illustrated below, GRPO often generates overly lengthy, step-by-step narratives. It may repeatedly restate the problem or elaborate on irrelevant details, even when the required operation is trivial.
In contrast, \ours{} typically produces a concise, task-sufficient reasoning trace that directly targets the required computation.

\modelquestion{
Danny collects bottle caps and wrappers. He found 65 wrappers and 5 bottle caps at the park. Now he has 31 bottle caps and 9 wrappers in his collection. How many more wrappers than bottle caps did danny find at the park?
}
\modelgt{
\boxed{60}
}
\modelresponseone{
Alright, I have this problem about Danny who collects bottle caps and wrappers. I need to figure out how many more wrappers than bottle caps he found at the park. Let's break it down step by step.

\#\#\# Understanding the Problem

First, I need to understand what the problem is asking. Danny has a collection of bottle caps and wrappers. He found some at the park, and now he has some in his collection. The problem gives me:

- **Found at the park:**
  - Wrappers: 65
  - Bottle caps: 5

- **Now in his collection:**
  - Bottle caps: 31
  - Wrappers: 9

The question is: **How many more wrappers than bottle caps did Danny find at the park?**

Wait, that seems a bit confusing. Let's rephrase it to make sure I understand.

Danny found 65 wrappers and 5 bottle caps at the park. Now, in his collection, he has 31 bottle caps and 9 wrappers. The question is asking for the difference between the number of wrappers he found at the park and the number of bottle caps he found at the park.

But looking at the numbers:

- Found at the park:
  - Wrappers: 65
  - Bottle caps: 5

So, the difference would be 65 wrappers - 5 bottle caps = 60.

But the problem mentions his current collection, which has 31 bottle caps and 9 wrappers. I'm not sure if that's relevant to finding out how many more wrappers than bottle caps he found at the park.

Wait, maybe I need to consider that the 31 bottle caps and 9 wrappers include both what he found at the park and what he already had before. But the problem is specifically asking about what he found at the park.

Let's think carefully.

\#\#\# Breaking Down the Information

Let's list out the information:

1. **Found at the park:**
   - Wrappers: 65
   - Bottle caps: 5

2. **Now in his collection:**
   - Bottle caps: 31
   - Wrappers: 9

The question is: **How many more wrappers than bottle caps did Danny find at the park?**

So, it's asking for the difference between the number of wrappers found at the park and the number of bottle caps found at the park.

Given that:

- Wrappers found at the park: 65
- Bottle caps found at the park: 5

The difference would be:

65 wrappers - 5 bottle caps = 60

But I need to make sure that this is the correct interpretation.

\#\#\# Considering the Current Collection

The problem also provides the current number of bottle caps and wrappers in Danny's collection:

- Bottle caps: 31
- Wrappers: 9

This might be relevant if we need to find out how many he had before finding the new ones at the park. But the question is only about what he found at the park, not about his total collection.

So, perhaps the information about his current collection is extra information, and we don't need it to answer the question.

\#\#\# Double-Checking the Interpretation

Let's rephrase the question to ensure I understand it correctly:

"How many more wrappers than bottle caps did Danny find at the park?"

This is asking for the difference between the number of wrappers he found at the park and the number of bottle caps he found at the park.

Given:

- Wrappers found at the park: 65
- Bottle caps found at the park: 5

So, the difference is:

65 - 5 = 60

Therefore, Danny found 60 more wrappers than bottle caps at the park.

\#\#\# Final Answer

After carefully analyzing the problem, the number of more wrappers than bottle caps Danny found at the park is:

\boxed{60}

}

\modelresponsetwo{
To determine how many more wrappers than bottle caps Danny found at the park, we need to follow these steps:

1. Identify the number of wrappers Danny found at the park:
   \[
   \text{Wrappers found} = 65
   \]

2. Identify the number of bottle caps Danny found at the park:
   \[
   \text{Bottle caps found} = 5
   \]

3. Calculate the difference between the number of wrappers and bottle caps found:
   \[
   \text{Difference} = \text{Wrappers found} - \text{Bottle caps found} = 65 - 5 = 60
   \]

Therefore, Danny found \boxed{60} more wrappers than bottle caps at the park.
}

These examples illustrate that \ours{} reduces overthinking primarily by trimming redundant problem restatement and unproductive reasoning on easy inputs, while still maintaining a clear solution trace.
This behavior aligns with our easy-side regularization, which discourages verbosity when additional reasoning is unlikely to improve correctness.

\section{Details of Reflective Words}
\label{sec:reflective_words}
Following the methodology described in Section~\ref{subsec:reasoning_pattern}, we use a curated set of \num{17} reflective words and phrases to identify instances of backtracking, self-correction, and verification in the model's reasoning traces.
A response is classified as \textit{reflective} if it contains at least one of the following terms:
\begin{itemize}
\item \textbf{Prefix-based terms:}
\textit{re-check}, \textit{re-evaluate}, \textit{re-examine}, \textit{re-think}, \textit{recheck}, \textit{reevaluate}, \textit{reexamine}, \textit{reevaluation}, \textit{rethink}.

\item \textbf{Verification and retry phrases:}
\textit{verify}, \textit{check again}, \textit{think again}, \textit{try again}, \textit{double-check}, \textit{double check}.

\item \textbf{Transition markers:}
\textit{wait}, \textit{yet}.
\end{itemize}

By monitoring these markers, we can distinguish between productive and redundant reasoning, where the latter generates lengthy responses without productive reflection.

\section{Limitations and Future Work}

\paragraph{Improving difficulty estimation.}
Like other group-based RL methods such as GRPO, \ours{} relies on a group of \(G\) rollouts per prompt to estimate advantages and derive a difficulty signal. 
While this proxy is lightweight in practice and adapts naturally during training, it can be noisy when the rollout group size \(G\) is small or when reward signals are sparse in certain domains. 
As a result, developing more robust difficulty estimators (e.g., uncertainty-aware signals) could further improve compute allocation for adaptive reasoning.

\paragraph{Scaling to longer contexts and larger models.}
Due to resource constraints, we do not systematically scale \ours{} to larger backbones or substantially longer context lengths.
For example, in the extreme \textit{Hard}-training setting in Section~\ref{subsec:robustness_to_training_difficulty_shifts}, we observed that a \(16K\) maximum response length resulted in a clipping ratio exceeding \(5\%\) for extremely difficult instances, compared to less than \(2\%\) in our main experiments.
This higher clipping ratio suggests that some productive reasoning trajectories can be prematurely truncated, potentially introducing additional optimization noise.
Consequently, the current context-length budget and training-set composition may not fully unlock the model’s potential under more generous compute budgets.
However, despite these constraints, systematic experiments across multiple benchmarks and model scales (4B/8B/14B) show that \ours{} consistently achieves genuine adaptiveness: \ours{} reallocates compute to trim overthinking on simple queries while preserving the complex, reflective reasoning patterns required for harder problems.
We leave systematic scaling studies to longer contexts and larger backbones for future work.


\end{document}